\documentclass[11pt, a4paper, logo, onecolumn, copyright]{googledeepmind} %
\pdfoutput=1
\usepackage[subtle]{savetrees}

\usepackage{wrapfig}
\usepackage{listings}
\usepackage[sort&compress,numbers]{natbib}%
\usepackage{xspace}
\usepackage{setspace}

\usepackage[dvipsnames]{xcolor}

\title{SpatialVLM: Endowing Vision-Language Models with Spatial Reasoning Capabilities}

\author{\textbf{\normalsize{Boyuan Chen$^{*, \dagger, 1}$, Zhuo Xu$^{*, 1}$, Sean Kirmani$^{1}$, Danny Driess$^{1}$, Pete Florence$^{1}$}} 
\\\textbf{\normalsize{Brian Ichter$^{1}$, Dorsa Sadigh$^{1}$, Leonidas Guibas$^{2}$, Fei Xia$^{1}$}}\\
$^1$Google DeepMind, $^2$Google Research \\
Correspond to: \texttt{boyuanc@mit.edu, zhuoxu@google.com, xiafei@google.com}\\
Website: \href{https://spatial-vlm.github.io/}{https://spatial-vlm.github.io/}
}

\newcommand{\algo}{SpatialVLM\xspace}

\begin{abstract}
     \includegraphics[width=1.0\linewidth]{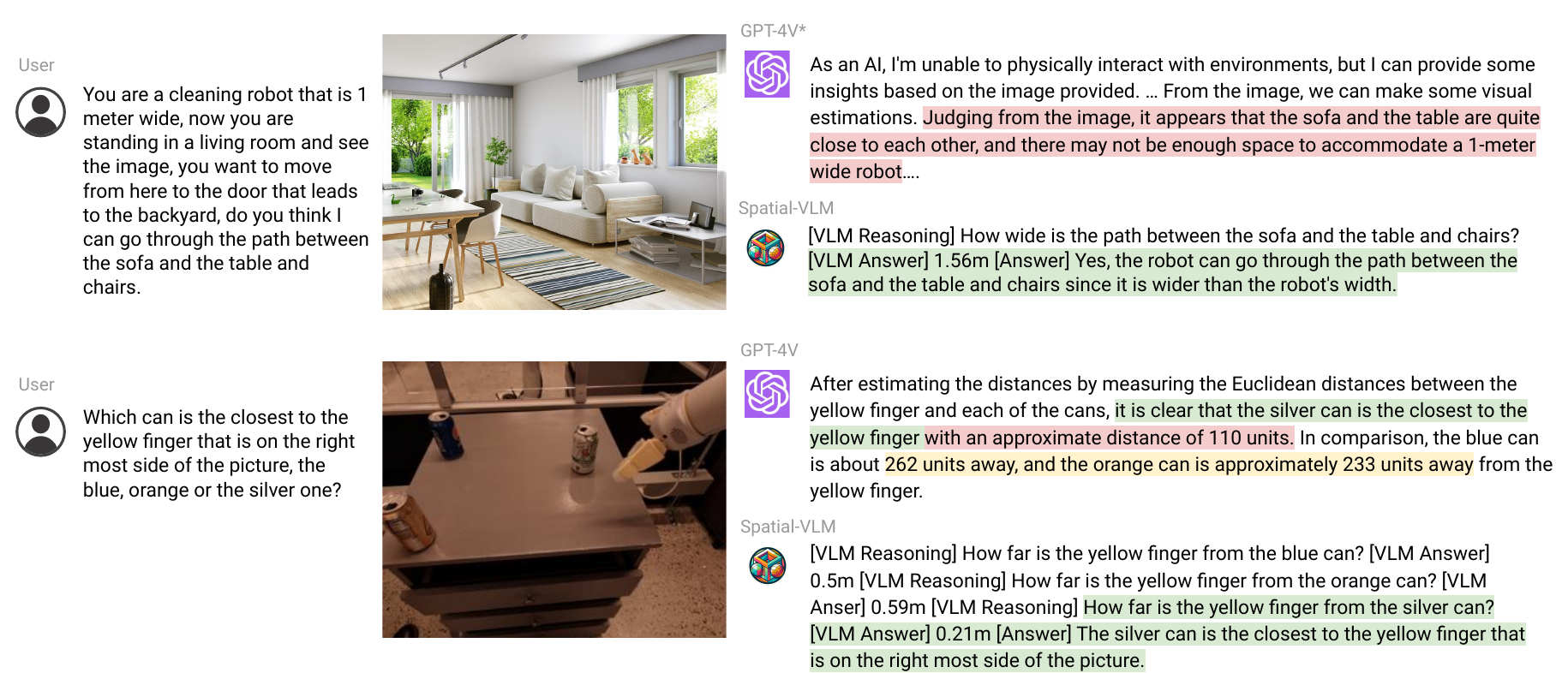}
    \captionof{figure}{We present \textbf{\algo}, a data synthesis and pre-training mechanism to enhance VLMs' spatial reasoning capabilities. We demonstrate that VLMs trained on our synthetic data exhibit strong spatial reasoning capabilities, and can generate metric distance estimation from 2D input images, addressing blind spots of current state-of-the-art VLMs like GPT-4V. ($^*$ GPT-4V accessed Nov. 2023).}
    \label{fig:teaser}
    \vspace{2em}

Understanding and reasoning about spatial relationships is a fundamental capability for Visual Question Answering (VQA) and robotics. While Vision Language Models (VLM) have demonstrated remarkable performance in certain VQA benchmarks, they still lack capabilities in 3D spatial reasoning, such as recognizing quantitative relationships of physical objects like distances or size difference. We hypothesize that VLMs' limited spatial reasoning capability is due to the lack of 3D spatial knowledge in training data and aim to solve this problem by training VLMs with Internet-scale spatial reasoning data. To this end, we present a system to facilitate this approach. We first develop an automatic 3D spatial VQA data generation framework that scales up to 2 billion VQA examples on 10 million real-world images. We then investigate various factors in training recipe including data quality, training pipeline and VLM architecture. Our work features the first Internet-scale 3D spatial reasoning dataset in metric space. By training a VLM on such data, we significantly enhance its ability on both qualitative and quantitative spatial VQA. Finally, we demonstrate that this VLM unlocks novel downstream applications in chain-of-thought spatial reasoning and robotics due to its quantitative estimation capability. 
\end{abstract}

\begin{document}
\maketitle

\section{Introduction}
Vision language models (VLMs) have made significant progress in recent years across a variety of tasks including image captioning, visual question answering (VQA), embodied planning, action recognition, and more \cite{alayrac2022flamingo, driess2023palme, gan2022vision, hu2022scaling}. While VLMs are powerful general-purpose models for a wide range of tasks, most state-of-the-art VLMs still struggle with \emph{spatial} reasoning, i.e.\ tasks that require understanding the position of objects in 3D space, or spatial relationships between them. Spatial reasoning capabilities are useful in their own right, but also for downstream applications such as in robotics or AR. For example, a spatial reasoning-imbued VLM can be used as a better general-purpose reward annotator~\cite{rocamonde2023vision} and success detector~\cite{du2023vision}.

The exploration of foundation models like VLMs is often inspired by human capabilities. Humans, through embodied experiences and evolutionary development, possess innate spatial reasoning skills. We effortlessly determine spatial relationships, such as the positioning of objects relative to each other or estimating distances and sizes, without complex chain-of-thoughts or mental computations. This natural proficiency in direct spatial reasoning tasks contrasts with the current limitations of VLMs and thus prevents them from accomplishing real-world tasks that requires multiple steps of spatial reasoning. This gap leads us to a compelling research question: can we imbue VLMs with spatial reasoning abilities akin to those of humans?

Therefore, we hypothesize that the limited the spatial reasoning abilities of current VLMs is not due to a fundamental limitation of their architecture, but rather is a limitation in common datasets available at scale on which such models are trained. For example, many VLMs~\cite{driess2023palme, chen2022pali, liu2022visual} are trained on internet-scale datasets characterized by image-caption pairs~\cite{chen2015microsoft}, which contain limited spatial information. This is partially due to the difficulties of obtaining spatial-information-rich embodied data or high-quality human annotations for 3D-aware queries.

Automatic data generation and augmentation techniques are one approach to deal with the data limitation problem \cite{richter2016playing, savva2019habitat, xia2018gibson, kolve2017ai2}. However, most previous data generation efforts focus on rendering photorealistic images with ground truth semantic annotation but overlook the richness of objects and 3D relationships. In contrast, we focus on extracting spatial information directly from real world data in order to capture the diversity and complexity of the true 3D world.

Our key insight is that recent advancement in off-the-shelf vision models can automatically generate rich 3D spatial annotations from 2D images. To this end, we propose a system called \algo that enables data generation and training of VLMs to enhance their spatial reasoning capabilities. Concretely, by combining 1) open-vocabulary detection, 2) metric depth estimation, 3) semantic segmentation and 4) object-centric captioning models, we can densely annotates real world data at scale. \algo converts the data generated by vision models into a format can be used to train VLMs on a mixture of captioning, VQA and spatial reasoning data. 

Through experiments, we find our trained VLM exhibit many desirable capabilities. First, its ability to answer qualitative spatial questions is greatly enhanced. Secondly, it can perform quantitative estimation reliably despite noisy training data. Such capability not only gives it common sense knowledge about object sizes but also makes it useful as a open-vocabulary reward annotator for rearrangement tasks. Thirdly, we find this spatial Vision Language Model, benefiting from its natural language interface, can perform spatial chain-of-thought to solve complex spatial reasoning tasks when combined with a powerful Large Language Model.

Our main contributions are:
\begin{itemize}
    \item We endow VLMs quantitative spatial reasoning capability, which is a fundamental capability of humans.
    \item We design a framework to automatically label 3D spatial reasoning VQA data based on real world images at the Internet scale.
    \item We study various training recipes: data quality, training pipeline, freeze/unfreeze visual encoder, etc, and investigate how they affect the learning quality.
    \item We demonstrate new capabilities of \algo in complex reasoning and robotics unlocked by the introduced task and method.
\end{itemize}

\section{Related Work}

\paragraph{Learning Spatial Reasoning.} Spatial distance estimation has been traditionally addressed as a part of broader tasks, such as SLAM~\cite{durrant2006simultaneous,cadena2016past} or depth estimation~\cite{fu2018deep}. 
When applying these spatial concepts to reasoning, prior works often focus on explicit spatial scene memories~\cite{gordon2018iqa, gervet2023navigating} or spatial scene graphs \cite{wald2020ssg, hildebr2020scene, walter2013learning, hemachandra2014learning}. 
Scene graphs allow interpretable, structured, statistical relation learning based on the spatial structures they encode. 
To answer spatial problems in VQA formats, they must handle it explicitly as a pathfinding problem on said scene graph. 
VLMs, on the other hand, are pretrained on large amounts of loosely structured information from vision-language datasets. Unlike scene graphs, the spatial understanding is encoded \emph{implicitly}. We can infuse the depth and 3D structure into the weights with an auxiliary task \cite{khansari2022practical,liu2023prismer}, capturing the relational information. In our work, we address the spatial relationship problem directly in the VLM, without an explicit underlying scene graph. In addition to understanding relative relationships in qualitative terms, we also explore estimating explicit metric distance relationships between objects in a scene.

\paragraph{Grounding Vision-Language Models.} Large language models (LLMs) are trained on internet-scale data, making them effective commonsense reasoners. 
However, LLMs (and by extension VLMs) may lack the necessary grounding to perform well at social reasoning \cite{kwon2023grounded}, physical reasoning \cite{gao2023physically}, physics reasoning~\cite{liu2022mind}, embodied tasks~\cite{huang2022language,shridhar2020alfred,ahn2022can}, and spatial reasoning tasks \cite{rozanova2021grounding, liu2022visual}.
Though language model with interactive world experience show grounding improvements~\cite{zellers2021piglet,xiang2023language}, the introduction of large vision models, such as Flamingo \cite{alayrac2022flamingo}, PaLI \cite{chen2022pali}, or PaLM-E \cite{driess2023palme}, has enabled a leap in performance.
These visually-grounded models have been used for several downstream tasks, such as in robotic success detection \cite{du2023visionlanguage,driess2023palme,sermanet2023robovqa,xiao2022robotic}, action prediction~\cite{shridhar2022cliport,brohan2023rt}, and reward prediction~\cite{fan2022minedojo,nair2022learning,cui2022can,mahmoudieh2022zero}. 
In this work we approach the problem of spatial reasoning through finetuning a VLM on a generated VQA dataset.
By directly finetuning a VLM on this task, we inherit the generality and reasoning capabilities of the underlying VLM as well as show how this approach is capable of tasks like reward generation.

\paragraph{Spatial Information in Vision-Language Datasets.} Many prior works have focused on benchmarking VLMs~\cite{xu2023lvlm,thrush2022winoground}, considering tasks like VQA (e.g. VQAv2 \cite{goyal2017making}, OK-VQA \cite{marino2019okvqa}, COCO \cite{lin2015microsoft}, or Visual Genome \cite{krishna2016visual}). 
Others have focused on fine-grained scene understanding, such as semantic segmentation~\cite{balavzevic2023towards, kirillov2023segment}, object detection~\cite{chen2021pix2seq}, or object identification~\cite{thomason2022language,cohen2019grounding}.
Others have focused specifically on spatial reasoning as a task, answering questions about object spatial relations (e.g., above, below, left, right) in real \cite{rozanova2021grounding, liu2022visual} or simulated~\cite{johnson2016clevr} scenes.
Real data in this domain can be limited by the amount generated by human labelers, while synthetic data has inherently bounded expressivity. 
In this work we consider how to automatically generate real data, and focus on the problem of not just spatial relations, but metric spatial distances, which can be directly applied to many downstream tasks.

\section{SpatialVLM}
\label{sec:method}
To equip VLMs with both qualitatively and quantitatively spatial reasoning capabilities, we propose to generate a large-scale spatial VQA dataset, which is used to train VLMs. Concretely, we design a comprehensive data generation framework which first leverages off-the-shelf computer vision models including open-vocabulary detection, metric depth estimation, semantic segmentation and object-centric captioning models to extract object-centric contexts, and then adopts template-based approach to generate massive spatial VQA data of reasonable quality. We train our SpatialVLM using the generated dataset to learn direct spatial reasoning capabilities, which we can then combine with the high-level commonsense reasoning embedded in LLMs to unlock chain-of-thoughts spatial reasoning.

\subsection{Spatial Grounding from 2D Images}
\begin{figure*}[t]
  \centering
  \includegraphics[width=0.95\linewidth]{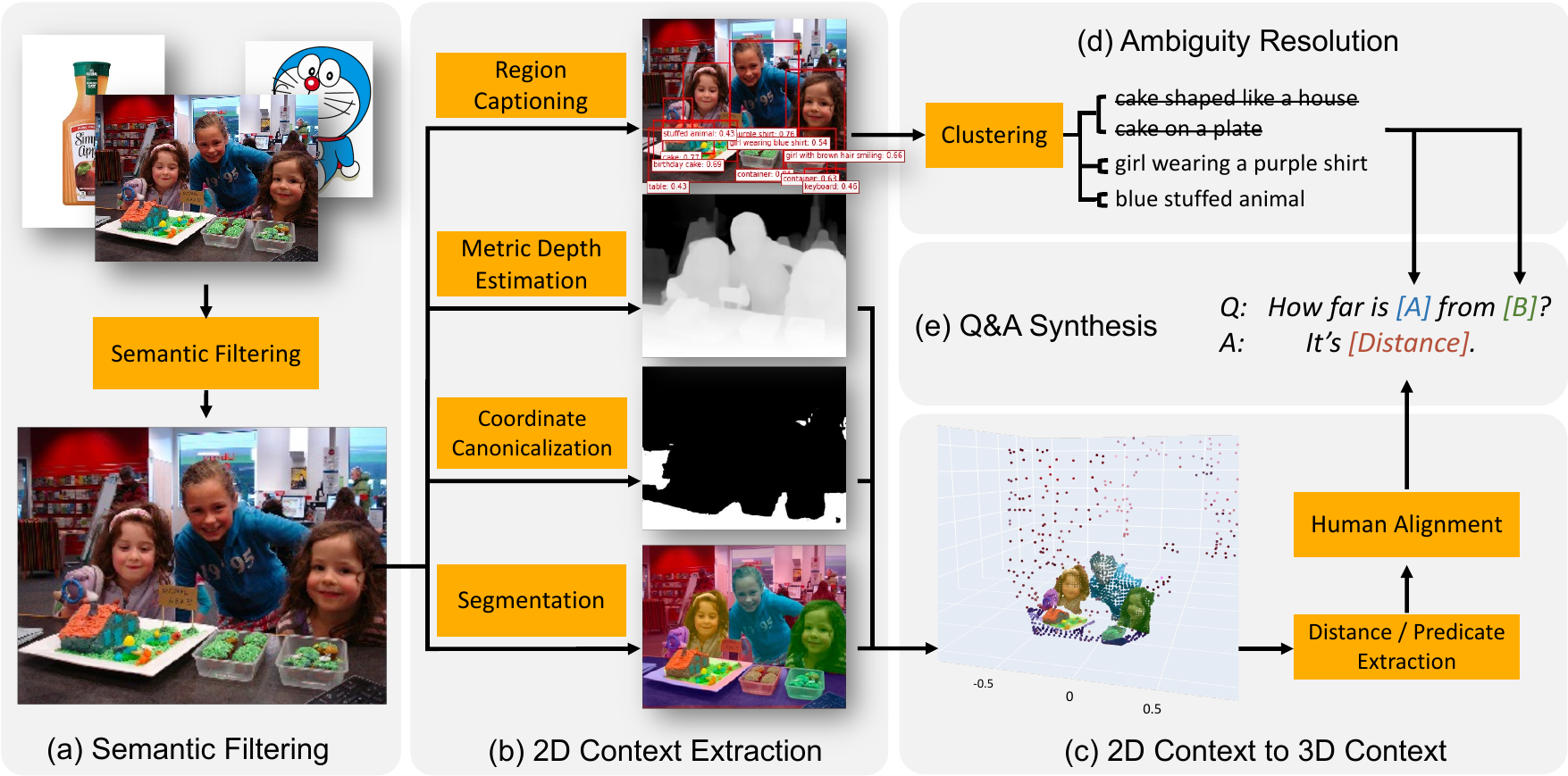}
  \caption{\textbf{An overview of our data synthesis pipeline.} (a) We use CLIP to filter noisy internet images and only keep scene-level photos. (b) We apply pre-trained expert models on internet-scale images so that we get object-centric segmentation, depth and caption. (c) We lift the 2D image into 3D point clouds, which can be parsed by shape analysis rules to extract useful properties like 3D bounding box. (d) We avoid asking ambiguous questions by clustering object captions using CLIP similarity score (e) We synthesize millions of spatial question and answers from object captions and extracted properties. }
  \label{fig:method}
\end{figure*}

We hypothesize that the reason for the lack of spatial reasoning capabilities of today's VLMs is not their architecture, but the lack of spatial reasoning training data. Following this insight, we design a pipeline that generates VQA data containing spatial reasoning questions. The pipeline is summarized in in Figure \ref{fig:method} and described in detail as follows.

\paragraph{Semantic Filtering}
While internet-scale image-captioning datasets have been widely used in VLM training \cite{chen2022pali}, many images in these datasets are not suitable for synthesizing spatial reasoning QA, due to the fact that they either consist of a single object or don't have a scene background (e.g. product pictures on shopping websites or screenshots of computer screen). Therefore, as the first step in our data synthesis pipeline, we adopt a CLIP-based open-vocabulary classification model to classify all images and rule out those that are not suitable.


\paragraph{Object-centric Contexts Extraction from 2D Images}
In order to extract object-centric spatial contexts from 2D images, we leverage a series of off-the-shelf expert models, including region proposal, region captioning~\cite{anonymous2023flexcap}, and semantic segmentation~\cite{kuo2019shapemask} modules to extract object-centric information. With this step, we obtain object-centric entities consisting of pixel clusters as well as open-vocabulary caption descriptions.

\paragraph{Lifting 2D Contexts to 3D Contexts}
Traditional spatial VQA datasets generated using object detection and bounding box positioning~\cite{krishna2017visual} are limited to the 2D image plane (lack of depth or altitude contexts) and pixel-level reasoning (lack of metric-scale size and distance contexts). We perform depth estimation~\cite{bhat2023zoedepth} to lift monocular 2D pixels to metric-scale 3D point clouds. We further canonicalize the camera coordinate system of the point cloud into a geodetic coordinate system, which is done by horizontal surface (e.g. ``floor'', ``table top'') segmentation~\cite{chen2017deeplab} and frame transfer. To the best of our knowledge, we are the first to lift internet-scale images to object-centric 3D point clouds and use it to synthesize VQA data embedded with 3D spatial reasoning supervision.

\paragraph{Ambiguity Resolution}
Sometimes there are multiple objects of similar categories in one image, leading to ambiguities of their caption labels. For example, one same caption label ``cake'' can refer to multiple different cakes in a same image. Therefore, before we can ask questions about these objects, we need to make sure the reference expressions are not ambiguous. We made two key design choices that have been validated empirically to be effective in tackling this challenge:

\begin{itemize}

\item We deliberately choose to avoid common object detectors, which tend to produce fixed and coarse categories such as ``cake'', and adopt FlexCap~\cite{anonymous2023flexcap}, a user-configurable object-centric captioning approach. In practice, for each object we can sample a random caption of a variable length between $1-6$ words. As a result, our object annotations are fine-grained, such as ``cake shaped like a house'' and ``cup cake in plastic container''

\item We design a semantic-oriented post-processing algorithm that further remove ambiguities by augmenting or rejecting object captions. Details of this algorithm are shown in Appendix~\ref{appendix:ambiguity} . 

\end{itemize}

\subsection{Large-Scale Spatial Reasoning VQA Dataset}
\label{subsec:vqa_dataset}
As motivated in Section \ref{sec:method}, we focus our study on infusing ``straightforward'' spatial reasoning capabilities into VLMs by pretraining with synthetic data. Therefore, we synthesize spatial-reasoning QA pairs that involve no more than two objects (denoted ``A'' and ``B'') in the image and consider the two following categories of questions.

\paragraph{Qualitative questions:} those that ask for judgement of some spatial relations. Examples are ``Given two objects A and B, which is more towards the left?'', ``Is object A more elevated than object B?'' and ``Among A and B, which is bigger in width?''.

\paragraph{Quantitative questions:} those that ask for more fine-grained answers that include numbers and units. Examples include ``how much to the left is object A compared to object B?'', ``How far is object A from the B?'', ``Find out how far A is positioned behind B relative to the camera.''.
Similar to the aforementioned examples, such questions can be synthesized using a main question template, and one can fill the object name entries using the object captions after disambiguation. This property allows us to do template-based generation, an approach commonly adopted by instruction tuning works~\cite{wei2021finetuned}. The answers to the questions are obtained through appropriate functions that we develop, which take as input the segmented point clouds and 3D bounding boxes of the relevant objects.

\begin{figure*}[t]
  \centering
  \includegraphics[width=0.95\linewidth]{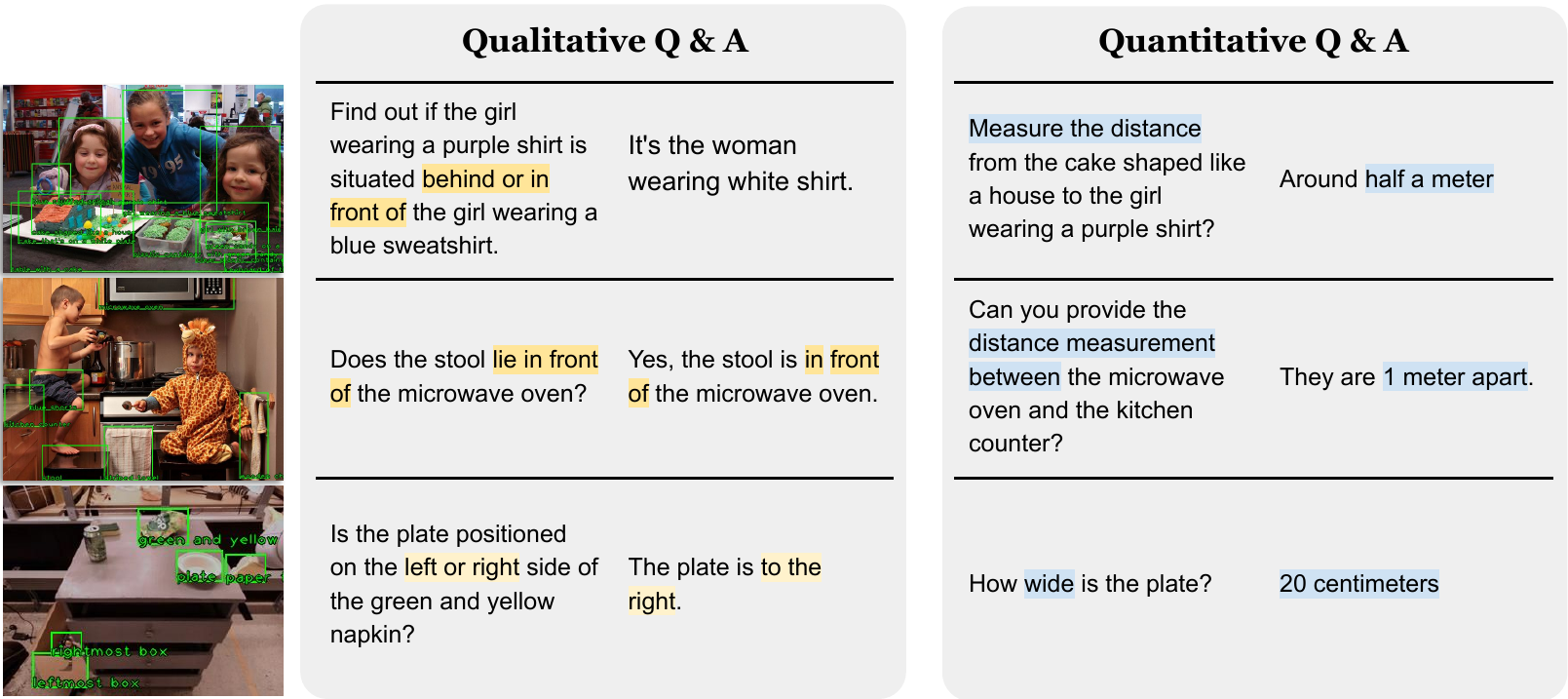}
  \caption{\textbf{Example data entries from the synthetic dataset}. Given the output of vision expert models, we follow a set of question generation template to generate both quantitative and qualitative question-answer pairs to highlight the diversity of the dataset. The spatial concepts are highlighted in blue. Such visual question-answer pairs can be easily mixed together with other captioning or question answering datasets and use the same training objectives.}
  \label{fig:s3vlm_dataset}
\end{figure*}



We designate $38$ different types of qualitative and quantitative spatial reasoning questions, each featuring around $20$ question templates and $10$ answer templates (we show examples in Appendix.~\ref{appendix:vqa_template}). We also add bias the sampling to encourage concise answers. Finally we introduce a human-aligned rounding mechanism in Appendix~\ref{appendix:alignment} to make number roundings in a human-like way. Using such an approach, we are able to generate ample question answering data pairs for the monocular camera images in webli and vqa datasets. Fig~\ref{fig:s3vlm_dataset} shows several example synthetic question answering pairs we obtained. In total, we create a massive dataset with $10$ million images and $2$ billion direct spatial reasoning QA pairs, featuring $50\%$ qualitative questions and $50\%$ quantitative questions. Thanks to the diversity of object captions and distance units, our synthetic dataset features significant diversity in terms of object description, question type and phrasing.

\subsection{Learning Spatial Reasoning}
\paragraph{Direct Spatial Reasoning} is defined as following, a Vision-Language Model takes as input an image $\mathcal{I}$ and a query $\mathcal{Q}$ of a spatial task, and output an answer $\mathcal{A}$, in the format of a text string, without using external tools or interacting with other large models. We adopt the same architecture and training procedure of PaLM-E~\cite{driess2023palme} except replacing PaLM~\cite{chowdhery2022palm} backbone with PaLM 2-S~\cite{anil2023palm}, a smaller variant. We then train our model using a mixture of the original PaLM-E dataset and our dataset, with $5\%$ of tokens dedicated to spatial reasoning tasks. Similar to PaLM-E, our method has the ability to perform VQA as well as basic embodied planning when combined. The key difference is that it can answer spatial reasoning questions about both binary predicates and quantitative estimations.

\paragraph{Chain-of-Thought Spatial Reasoning}

Many real-world tasks require multiple steps of spatial reasoning. For example, to determine if object A can fit into object B, one would need to reason about sizes and constraints. Sometimes one would need to reason over grounded spatial concept (e.g. the counter in the image is 1 meter high) and common sense knowledge (so that a toddler cannot reach it). \algo provides a \emph{natural language} interface to query with grounded concepts, when combined with a powerful LLM, we can perform complex spatial reasoning.

\begin{figure*}[t]
  \centering
  \includegraphics[width=0.9\linewidth]{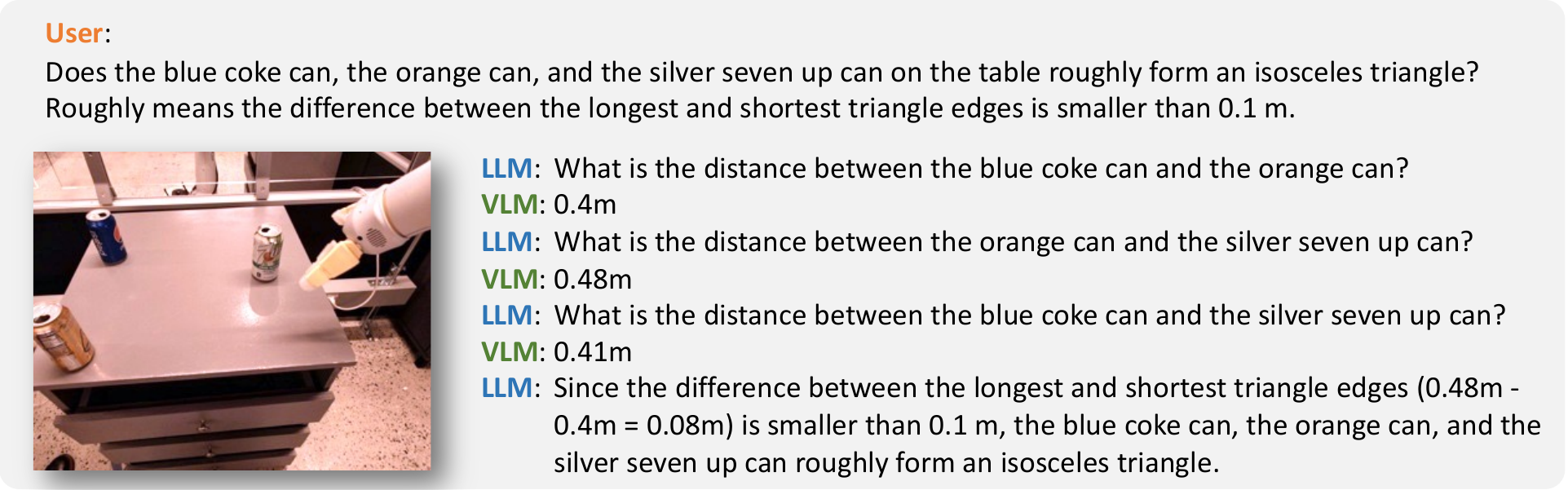}
  \caption{\textbf{Chain-of-thought spatial reasoning.} We illustrate that we can perform Chain-of-Thought Spatial reasoning with \algo. In this example, with the help of an LLM orchestrating \algo, the system is able to answer questions like ``Does the blue coke can, the red coke can, and the green sponge on the table roughly form an equilateral triangle".}
  \label{fig:s3vlm_cot}
\end{figure*}

We call this method \textbf{``Chain-of-Thought Spatial Reasoning"}. While our synthesized data only contains direct spatial reasoning questions, it's easy for a VLM to compose them together to solve complex questions that require multi-hop chain-of-thought reasoning. Similar to the method in Socratic Models~\cite{zeng2022socratic} and LLM as coordinator~\cite{chen2023language}, we utilize an LLM (\texttt{text-davinci-003}) to coordinate and communicate with our \algo to solve complex problems with Chain-of-Thought prompting~\cite{wei2022chain} as shown in Fig.~\ref{fig:s3vlm_cot}. The LLM can break down complex questions into simple questions, query the VLM, and put the reasoning together to derive the result.

\section{Experiments}

We conduct experiments to answer the following questions: 
\begin{itemize}[label={},leftmargin=*]
    \item \textbf{Q1} Does our spatial VQA data generation and training pipeline improve VLM's general spatial reasoning capabilities? And how well does it perform? 
    \item \textbf{Q2} How does the noisy synthetic spatial VQA data and different training strategies affect the learning performance? 
    \item \textbf{Q3} Does the VLM equipped with ``direct'' spatial reasoning capabilities unlock new capabilities such as chain-of-thought reasoning and embodied planning?
\end{itemize}

We train our model using a mixture of PaLM-E training set and our spatial VQA dataset. To verify whether VLM's limitation in spatial reasoning is a data problem, we choose the following state-of-the-art VLMs as baselines, all trained on mixtures in which semantic-captioning tasks occupy a heavy weight, and without our spatial VQA dataset.

\noindent \textbf{GPT-4V}\footnote{Accessed Nov 2023 via OpenAI API.} GPT-4V is a version of GPT-4~\cite{openai2023gpt4} that supports multimodal input, it achieves state-of-the-art performance in many vision-language tasks.

\noindent \textbf{PaLI}~\cite{chen2022pali}. An encoder-decoder VLM trained on multi-lingual corpora, it shows state-of-the-art performance on captioning and visual-question answering tasks. We used PaLI-X 55B variant in our experiments.

\noindent \textbf{PaLM-E}~\cite{driess2023palme}. A VLM trained on internet-scale vision, language, and vision-language data, as well as robotics data. It shows state-of-the-art performance in OKVQA benchmark, as well as being capable of robot planning tasks. We used PaLM-E 12B across our experiments.

\noindent \textbf{PaLM 2-E} The vanilla PaLM 2-E is an updated version of PaLM-E\cite{driess2023palme} with exact same training procedure but a more recent LLM backbone.  Due to the shared network architecture and training procedure with \algo, vanilla PaLM 2-E naturally serves as the baseline to study the effect of generated data. In the rest of the paper, unless specifically noted, PaLM 2-E corresponds to PaLM 2-S in terms of parameter count following the naming convention in PaLM 2 technical report~\cite{anil2023palm}.

\noindent Finally, we consider open source models like \textbf{LLaVA-1.5}~\cite{liu2023visual} and \textbf{InstructBLIP}~\cite{instructblip}.

\subsection{Spatial VQA performance}

\label{section:vqa_performance}
To stress-test the VLM's spatial reasoning capabilities, a spatial reasoning VQA benchmark with guaranteed performance grounding is required. However, there is not such a proper benchmark available in the literature. Therefore, we created a benchmark by having human annotators label a diverse set of ``direct'' qualitative and quantitative VQAs on a subset of WebLI images~\cite{chen2022pali}, which are unseen to all VLMs during the training phase. The benchmark questions and answers are diverse and freeform, following the synthetic data generation pattern described in Section \ref{subsec:vqa_dataset} (details in Appendix.~\ref{appendix:vqa_human}). We annotated $331$ qualitative spatial reasoning VQA pairs and $215$ quantitative spatial reasoning VQA pairs.

\paragraph{Qualitative Spatial VQA} For such questions, both the human annotated answers and VLM outputs are freeform natural language. Therefore, to evaluate the performance of the VLMs, we use human raters to determine if an answer is correct, and show the success rates of the VLMs in Table.~\ref{tab:spatial-qual}. It is shown that \algo is able to achieve significantly higher accuracy compared to all baselines that are not trained using the synthetic spatial VQA data, surpassing other vision-language models including GPT-4V. Among the baselines, the second best model is LLaVA-1.5, which might be caused by their use of bounding boxes and corresponding captions in visual instruction tuning. Anecdotally, we found LLaVA-1.5 performs well in 2D spatial relationship inference, but inferior to our models in 3D spatial reasoning. This experiment suggests that large and high-quality spatial reasoning data is key to spatial reasoning capabilities, which are not present in pretraining datasets of state-of-the-art VLMs.

\begin{table*}[t]
    \centering
    \resizebox{0.75\textwidth}{!}{%
    \begin{tabular}{cccccccc}
        \toprule
        Method & GPT-4V & LLaVA-1.5 & InstructBLIP &PaLI & PaLM-E & PaLM 2-E & Ours \\
        \midrule
        Accuracy &  $68.0\%$ & $71.3\%$  &  $60.4\%$ & $60.7\%$ & $50.2\%$ & $50.4\%$ & $\mathbf{75.2\%}$ \\
        \bottomrule
    \end{tabular}
    }
    \caption{\textbf{Accuracy of different VLMs on binary predicate prediction tasks}. Our proposed method  outperform baselines on binary predicate prediction tasks by a large margin owing to the addition of synthetic data.}
    \label{tab:spatial-qual}
\end{table*}

\begin{table*}[t]
    \centering
        \resizebox{0.8\textwidth}{!}{%

    \begin{tabular}{cccccccc}
        \toprule
         & GPT-4V & LLaVA-1.5 & InstructBLIP & PaLI & PaLM-E & PaLM 2-E & Ours \\
        \midrule
        \text{Output numbers \%} & 1.0\% &20.9\% &26.0\% & 52.0\% & 83.2\% & 88.8\% & \textbf{99.0\%} \\
        \text{In range [50, 200]\%}& 0.0\% &13.0\% &7.9\% & 5.3\% & 23.7\% & 33.9\% & \textbf{37.2\%} \\
        \bottomrule
    \end{tabular}
    }
    \caption{\textbf{Accuracy of different VLMs on quantitative questions about spatial relationship.} As can be seen from this table, first, our method outputs valid format more often (99.0\% of the time) than baseline methods. Second, our method outputs quantitative distance estimation that is closer to ground truth annotated by human more often than baseline methods.}
    \label{tab:spatial-quant}
\end{table*}

\paragraph{Quantitative Spatial VQA} For these questions, both human annotator answers and the VLM outputs are natural language descriptions of distance, height, elevation, etc, using their preferred units. We design two metrics for evaluating the performance of the VLM. First, we use the success rate of the VLM to produce a number to reflect if the VLM is able to understand the quantitative spatial reasoning question. Second, since the answer can range widely from centimeters to kilometers, we use percentages of the VLM answers that fall into half to twice of the ground truth value to represent how accurate the VLM's estimates are. The results are shown in Table.~\ref{tab:spatial-quant}, and it is shown that our model performs better on both metrics than baselines with large margins. We observed that baseline VLMs are reluctant to give answers consisting of numbers. For example, replying \textit{``No."} to questions like \textit{``Can you tell me the distance between ..."}. This is likely due the the distribution of the training data. Additionally, we find that state-of-the-art VLM GPT-4V often refrain from generating answers about distance in SI units with a disclaimer text \textit{``I'm sorry, but I cannot provide an exact distance as the image does not offer precise references for measurement.."}. Our approach \algo achieves significantly higher success rate than all baselines, achieving in-range results on almost half of the questions. This performance is remarkable given that the human annotations are noisy, and agreement among annotators are not often guaranteed (Appendix.~\ref{appendix:vqa_human}). To better understand our model's performance and limitations, we visualized the relative error against the ground truth value in Fig.~\ref{fig:quant_error} in the Appendix. We found that \algo does well on medium range scenes like those with objects $1-10$ meters from the camera. This coincides with the range where our monocular depth estimator~\cite{bhat2023zoedepth} reliably outputs metric accurate depth estimations, which indicates that our method inherits the biases and limitations from expert vision models in the data synthesis pipeline.

\begin{figure*}[t]
    \centering
    \includegraphics[width=0.95\linewidth]{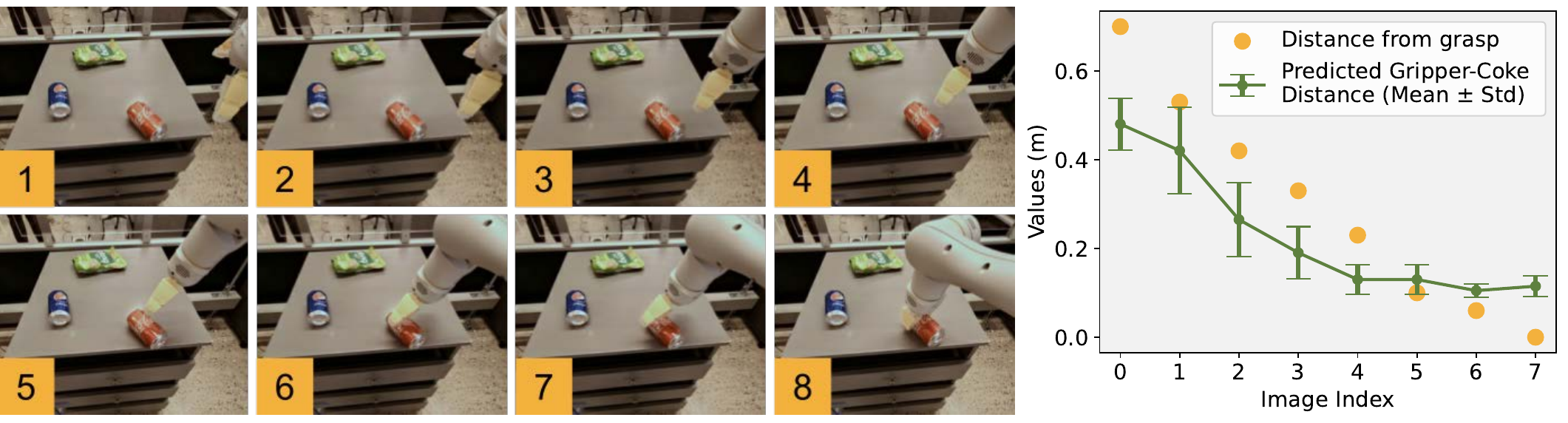}
    \caption{Given a sequence of images where the robot gripper is approaching the coke can, we ask \algo ``What is the distance between the yellow gripper and the coke can". We are able to get accurate and monotonically decreasing distance estimations.}
    \label{fig:fractal_qualitative}
\end{figure*}

\subsection{Effect of Spatial VQA Data to General VQA}

The second question we want to answer is: since we co-train with a considerable amount of spatial VQA data, whether the performance of VLM in other tasks will degrade as a result. We compared our model with the vanilla PaLM 2-E trained without the spatial VQA dataset on general VQA benchmarks, and as summarized in Table.~\ref{tab:vqa}, our model achieves comparable performance as PaLM 2-E on the OKVQA benchmark, in which limited spatial reasoning questions are included, and performs slightly better on VQA-v2 test-dev benchmark, which includes spatial reasoning questions. This seem to suggest that VLMs are generally underfitting in the distribution of tasks close to spatial reasoning, and can benefit from spatial VQA supervisions without hurting their general VQA capabilities.


\begin{table}[h]
    \centering
    \setlength{\tabcolsep}{8pt}
    \renewcommand{\arraystretch}{1.0}
    \resizebox{0.5\textwidth}{!}{%

    \begin{tabular}{ccc}
        \toprule
         General VQA benchmarks & OKVQA & VQA v2 \\
         \midrule
        PaLM 2-E w/o co-training & \textbf{61.4\%} & 76.6\%\\
        Ours & 61.0{\color{red}(-0.4)}\% & \textbf{79.0{\color{cyan}(+2.4)}\%}\\
        \bottomrule
    \end{tabular}
    }
    \caption{\textbf{VQA performance.} Co-training on \algo training mix and finetuning on VQA datasets (VQA v2) improves VQA performance. A PaLM 2-E model trained with \algo data improves VQA v2 performance by 2.4\% compared to 
    a model with the same number of parameters, but without the data. 
    However, we don't find OKVQA task to benefit from \algo training.}
    \label{tab:vqa}
\end{table}

\subsection{Effect of Visual Transformer (ViT) Encoder in Spatial Reasoning}

Does a frozen ViT (trained on contrastive objective) encode enough information to perform spatial reasoning?
To study this, we start at the 110k training step and branch into two training runs, 
one with the ViT frozen, the other with ViT unfrozen. We train both models for 70k steps, and evaluate percentages of answers from both models that fall into various ranges of the ground truth value in Table~\ref{tab:freeze_vit}.

\begin{table}[h]
    \centering
    \setlength{\tabcolsep}{8pt}
    \renewcommand{\arraystretch}{1.0}
    \resizebox{0.6\textwidth}{!}{%
    \begin{tabular}{cccc}
        \toprule
          & [50, 200]\% & [66.7, 150]\% & [90, 110]\% \\
         \midrule
        Frozen ViT & {34.9\%} & 9.3\% & 5.6\% \\
        Unfrozen ViT & \textbf{37.2{\color{cyan}(+2.3)}\%} & \textbf{10.7{\color{cyan}(+1.4)}\%} & \textbf{8.4{\color{cyan}(+2.8)}\%}\\
        \bottomrule
    \end{tabular}
    }
    \caption{Comparison on finetuning with frozen or unfrozen ViT. We find it is beneficial to unfreeze the pretrained ViT for distance estimation tasks.}
    \label{tab:freeze_vit}
\end{table}

It is shown that for larger scale and less fine-grained distance estimation, such as making a rough estimation with in the half-to-twice range of the ground truth, training without freezing ViT performs slightly worse but comparable with unfrozen ViT. However, for more fine-grained distance estimation like estimating accurate quantitative values, the model with unfrozen ViT performed considerably better. We hypothesize that the pretrained ViT (with contrastive or classification loss) is lossy in its fine-grained spatial information. Our model achieves $8.4\% $ accuracy for predicting a value $0.9\times$ to $1.1\times$ range of human annotation. This is remarkable since humans annotations are noisy. In fact, human sometimes tend to give noisy estimations, as they prefer to round an estimation of $0.8$ meter to $1$ meter. It remains challenging to evaluate quantitative spatial reasoning capabilities of vision-language models in broad domains.

\begin{table}[t]
    \centering
    \setlength{\tabcolsep}{8pt}
    \renewcommand{\arraystretch}{1.0}
    \resizebox{0.5\textwidth}{!}{%

    \begin{tabular}{ccccc}
        \toprule
         Gaussian std & 0 & 0.1 & 0.2 & 0.3 \\
         \midrule
        MSE(m) & 0.046 & 0.053 & 0.039 & 0.048 \\
        \text{[50, 200]\%} & 59.0\% & 55.8\% & 61.1\% & 61.1\% \\
        \bottomrule
    \end{tabular}
    }
    \caption{Comparison on different data noise levels, controlled using standard deviation (STD) of Gaussian noise. We find that our model can learn despite moderate amount of random noise.}
    \label{tab:fractal_noisy_data}
\end{table}

\subsection{Effect of Noisy Quantitative Spatial Answers}

Since the quantitative answers of the spatial VQA dataset are noisy, we study if VLMs can learn generalizable quantitative estimations from a large amount of noisy training data. To do so, we first come up with a domain where we are able to generate high quality quantitative answers. As discussed in Section~\ref{section:vqa_performance} the monocular depth estimation is one of the steps in the data generation pipeline that induce the most noises. Therefore, we leverage our robotic manipulation dataset, which provides near-ground-truth depth information captured using a depth camera. As a result, the generated quantitative answers are more accurate. We train VLM using this dataset, and find the model able to perform fine-grained distance estimation in the manipulation domain (Fig.~\ref{fig:fractal_qualitative}), which further demonstrates the data accuracy.

To study how noisy data affects VLM training, we add Gaussian noises upon the quantitative answers of the accurate manipulation spatial VQA dataset, and obtain a series of noisy datasets of different noise level. We train VLMs using the noisy datasets and evaluate them using a human annotated quantitative spatial VQA benchmark for manipulation. Table.~\ref{tab:fractal_noisy_data} compares how different Gaussian noise standard deviations affect the overall VLM performance on quantitative spatial VQA. Since the objects in the manipulation VQA datasets are within 1 meter range, we added the mean squared error (MSE) as a metric to evaluate the VLM performance, as well as the half-to-twice percentage which is defined in Section \ref{section:vqa_performance}. It is shown that VLMs trained on datasets of different noise levels achieve similar spatial reasoning accuracy. We speculate this is due to the noisy nature of the training data and the manually annotated evaluation benchmark, and that VLM can learn a spatial reasoning common-sense despite noisy data. We observed this interesting phenomenon in robotics experiments as well. In Fig.~\ref{fig:vlm_reward}, the distance estimation is exhibit a bias towards the mean since the model is heavily regularized.

\subsection{Spatial Reasoning Unlocks Novel Applications}

\begin{figure*}
    \centering
    \includegraphics[width=0.9\textwidth]{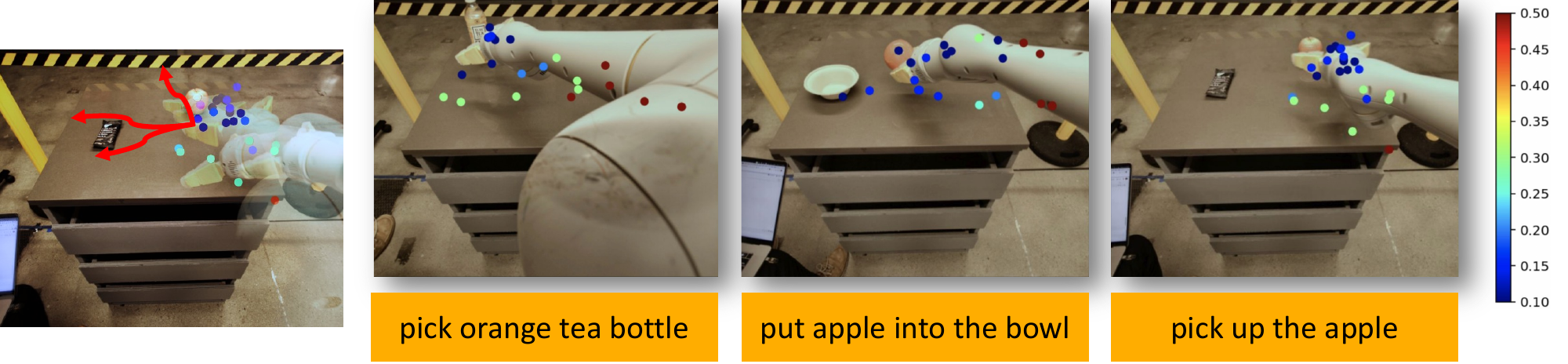}
    \caption{\small\textbf{\algo as reward generator for robotics tasks.} \algo provides a ``natural-language queriable" distance estimation tool, and can be used for robotics tasks. For example, for the task ``pick orange tea bottle", the reward/cost function can be the a function of the response of ``What is the distance between the yellow gripper fingers and the orange tea bottle". And for the task ``put the apple into the bowl", the reward/cost function can be a function of the response of ``what is the distance between the apple and bowl". We sample different gripper positions and show the cost function in the above scatter plots. }
    \label{fig:vlm_reward}
\end{figure*}

\paragraph{VLM as a Dense Reward Annotator} One important application of VLM is robotics. Recently, works have shown that VLMs and LLMs can serve as universal open-vocabulary reward annotators and success detector~\cite{du2023visionlanguage} for robotics tasks, which can be used to derive useful control policies. However, the reward annotation ability of VLMs are often limited by lack of spatial awareness. Since \algo is able to quantitatively estimate distances or sizes from image, it's uniquely suited as a dense reward annotator. We conduct a real robot experiment where we specify a task in nature language and ask \algo to annotate a reward for each frame in a trajectory. In Figure~\ref{fig:vlm_reward}, each dot illustrates an object location and their color indicates the annotated reward. As the robot makes progress towards the specified goal, we can see the reward increase monotonically, indicating the ability of \algo to serve as a dense reward annotator.

\paragraph{Chain-of-Thought Spatial Reasoning} In this section, we investigate whether \algo can be used to do tasks requiring multi-step reasoning, given its enhanced ability to answer elemental spatial questions. We demonstrate a few examples in Figure ~\ref{fig:teaser} and Figure ~\ref{fig:s3vlm_cot}. A large language model, in this case GPT-4, when equipped with \algo as a spatial reasoning submodule, can perform complex spatial reasoning tasks, such as answering if 3 objects in the environment can form a ``isosceles triangle".

\section{Conclusion}

In conclusion, our research addresses the challenge of infusing spatial reasoning to VLMs, and approach it by constructing a framework for automatic generation of 3D spatial reasoning VQA data based on Internet-scale real-world images. We ablate different design choices in the recipes for training VLMs, such as training with large amount of noisy data and unfreezing ViT. While our direct spatial queries are built on a finite set of templates, we show \algo can be extended to tackle more complicated chain-of-thought reasoning that requires spatial reasoning components. \algo is also demonstrated to be useful for robotics tasks, where we show that a 3D spatial-aware VLM could be used as a reward annotator for robotics tasks. Additional study of more nuanced geometric primitives can also help fully ground spatial reasoning in 3D geometry.

\bibliographystyle{plainnat}

\bibliography{main}

\newpage
\appendix
\section{Appendix}
\definecolor{lightgray}{gray}{0.95}

\subsection{Additional Experiments and Details}
\paragraph{Spatial VQA Human Annotated Benchmark}
\label{appendix:vqa_human}

\begin{figure*}[h]
  \centering
  \includegraphics[width=0.95\linewidth]{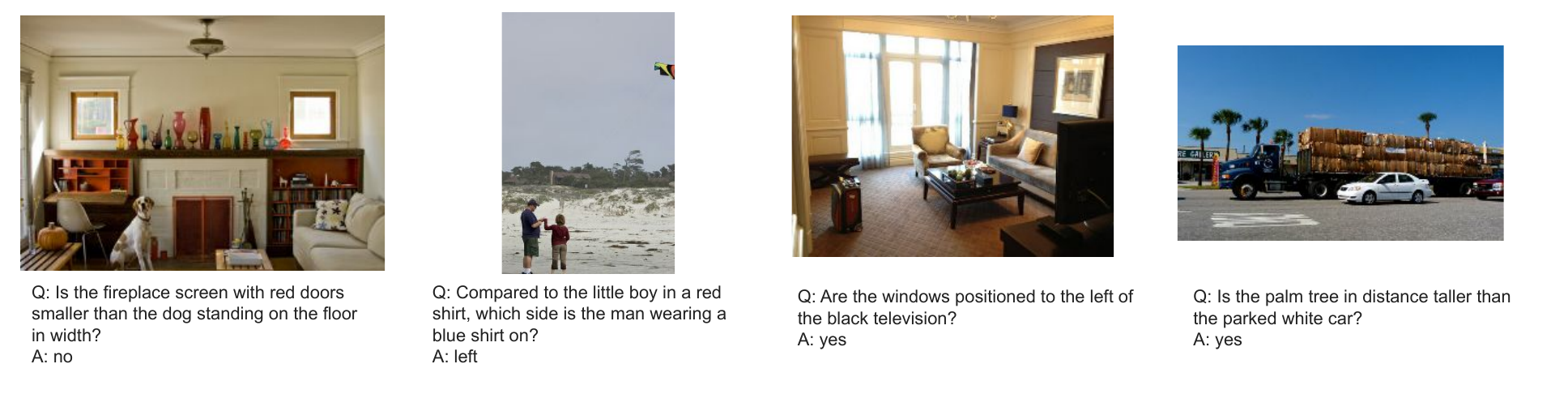}
  \caption{\textbf{Example question-answer pairs of the Spatial VQA qualitative benchmark}}
  \label{fig:qualitative_vqa_benchmark}
\end{figure*}

\begin{figure*}[h]
  \centering
  \includegraphics[width=0.95\linewidth]{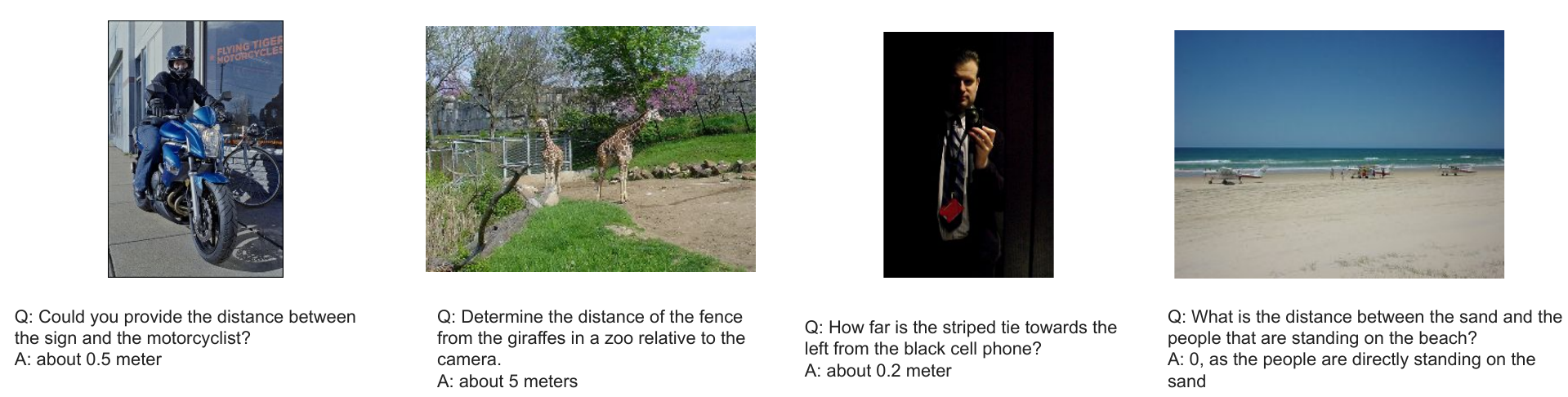}
  \caption{\textbf{Example question-answer pairs of the Spatial VQA quantitative benchmark}}
  \label{fig:quantitative_benchmark}
\end{figure*}

\begin{figure*}[h]
  \centering
  \includegraphics[width=0.95\linewidth]{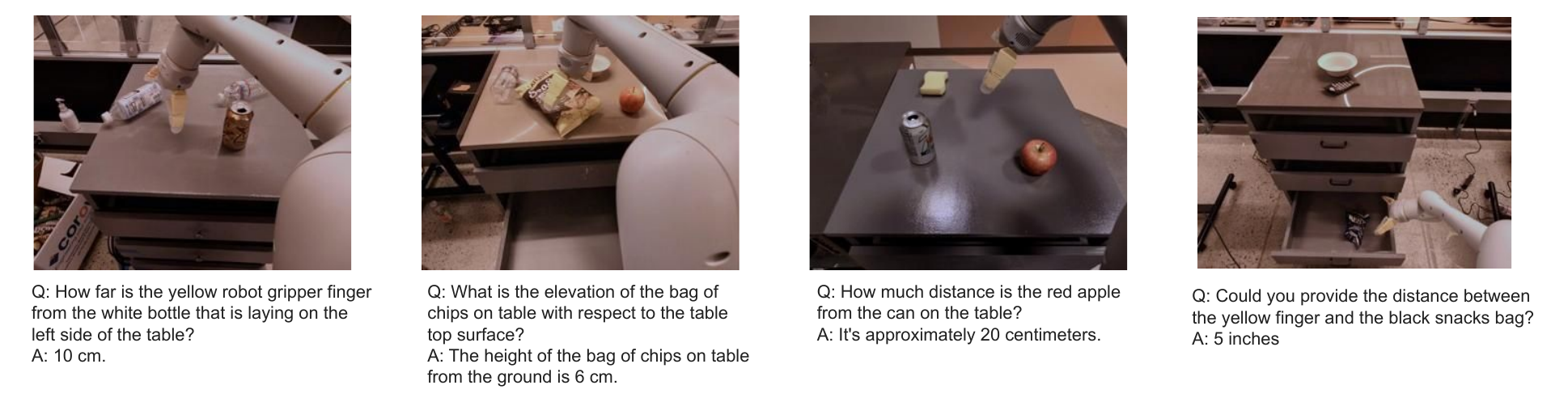}
  \caption{\textbf{Example question-answer pairs of the robotic manipulation VQA quantitative benchmark}}
  \label{fig:manipulation_quantitative_benchmark}
\end{figure*}

We manually labelled 546 qualitative and quantitative question pairs for WebLi and robotic manipulation VQA. In the human annotation pipeline, we use the spatial VQA data generation pipeline described in Section 3 to provide a sample question for each image, the human annotator would look at the image and the sample question, decide if he or she would like to use the question or type a more proper question, or to skip the annotation for the image. Then based on the sample question or the human input question, the human annotator would type the answer he or she thinks as proper in the form of natural language. Fig.~\ref{fig:qualitative_vqa_benchmark}, Fig.~\ref{fig:quantitative_benchmark}, and Fig.~\ref{fig:manipulation_quantitative_benchmark} shows examples of the human annotated spatial VQA pairs.

\begin{figure*}[t]
  \centering
  \includegraphics[width=0.95\linewidth]{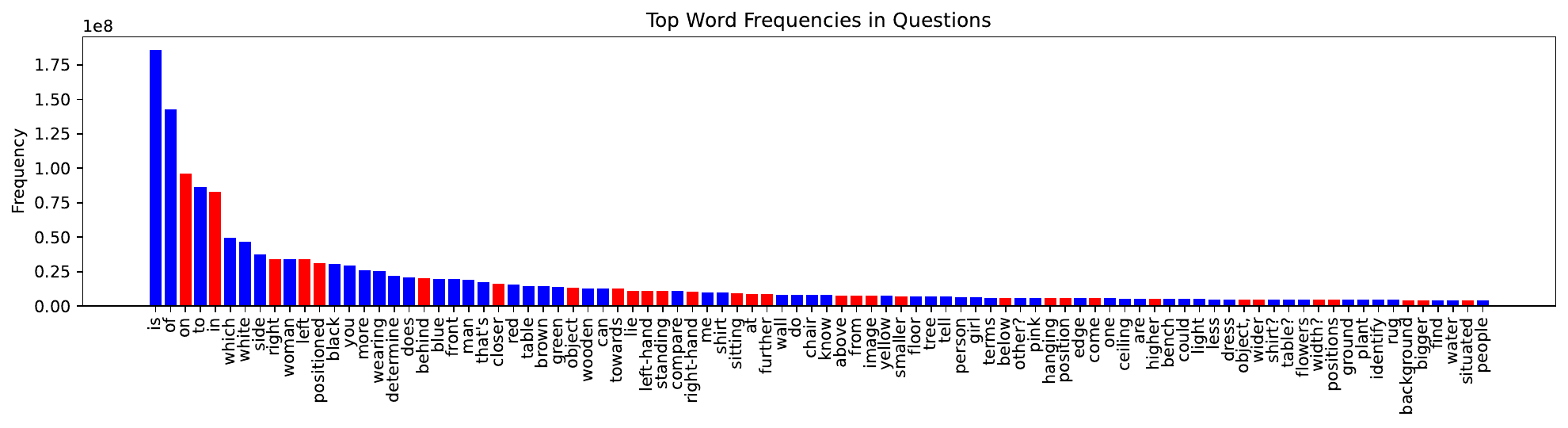}
  \caption{\textbf{Top word frequency}. Top words appeared in the training dataset, the red color indicate the word is involved in discussing a spatial concept. It shows that our training data is rich and diverse in spatial questions.}
  \label{fig:qualitative_benchmark}
\end{figure*}

\begin{figure}[h]
\centering
    \includegraphics[width=0.45\textwidth]{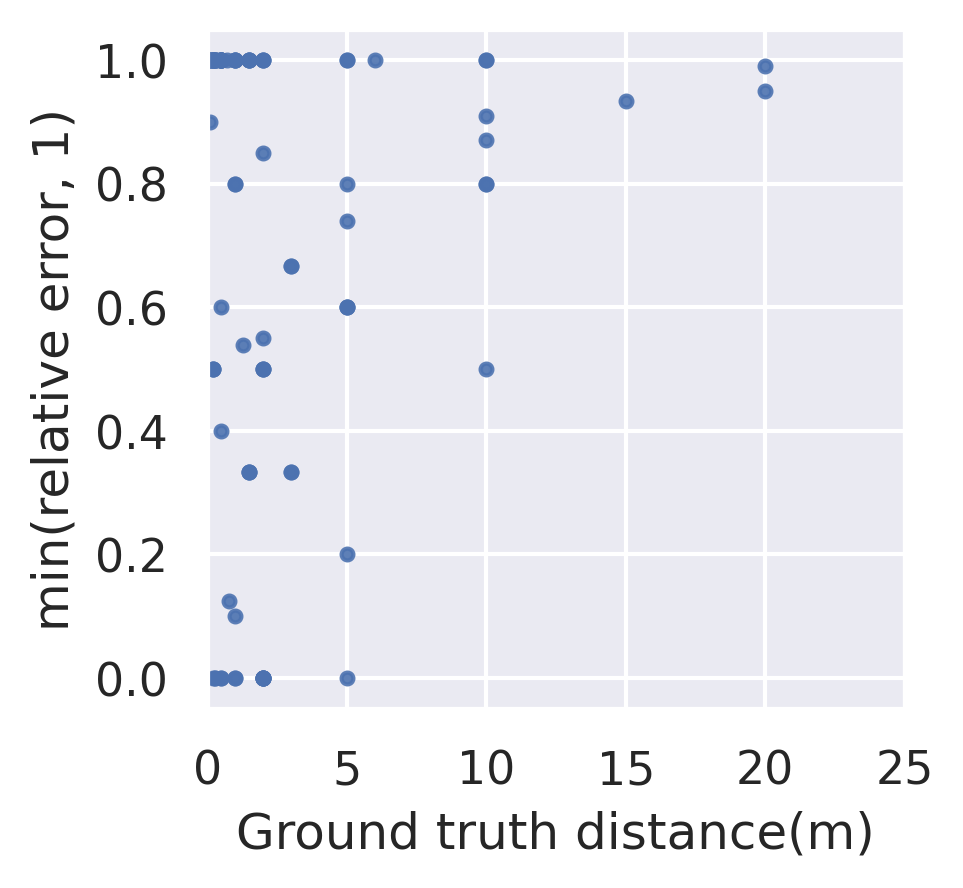}
    \caption{\textbf{Error vs Scene Depth ablation.} The errors that \algo make eventually attributes to the noise in the data, we plot the distance estimation relative error (capped at 1.0) w.r.t. ground truth distance, and found that there are generally larger errors for bigger distance. We hypothesize this might be due to dataset bias of ZoeDepth~\cite{bhat2023zoedepth}.}
    \label{fig:quant_error}
\end{figure}

\paragraph{Chain-of-thoughts}
Here we provide more details to our implementation of chain-of-thought spatial reasoning. As we mentioned in main paper, we prompt a LLM to perform chain-of-thought reasoning with ability to query our \algo for visual information. Since the LLM isn't aware of visual information itself, we prompt it to make decision as if it's playing a game, by asking its friend who can see an image that it cannot see itself. We provide the full prompt below:

\begin{lstlisting}[basicstyle=\ttfamily\scriptsize, backgroundcolor = \color{lightgray}, keywordstyle = {\textbf}, caption={\algo CoT Prompts}, label={lst:cot}]

You are participating in a visual question answering game with your friend. In this game, you are presented with a question which requires visual information from an image to answer. You can see the question but not the image, while your friend can see the image but not the original question. Luckily, you are allowed to decompose the question and ask your friend about the image. Your friend gives you answers which can be used to answer the original question.

Here is a sample conversation:
[Question] How can I clean up the table? Give detailed instruction about how should I move my hand.
[You] What objects are there in the image?
[Friend] There is an empty coke can, a trash bin and a coffee machine.
[You] Is the trash bin to the left or to the right of the coke can?
[Friend] It's to the left.
[You] Is the trash bin or the coke can further from you?
[Friend] They are similar in depth.
[You] How much to the left is the trash bin compared to the coke can?
[Friend] Around 20 centimeters.
[Answer] One should grab the coke can, move it 20 centimeters left and release it so it falls in the trash bin.

Here is another example:
[Question] Tell me if the distance between the blue bottle and the yellow book is longer than that between the plant and the coke can?
[You] What is the distance between the blue bottle and the yellow book?
[Tool] 0.3m
[You] What is the distance between the plant and the coke can?
[Friend] 0.7m
[Robot] Since the distance between the blue bottle and the yellow book is 0.3m and distance between the plant while the coke can is 0.7m, the distance between the blue bottle and the yellow book is not longer than that between the plant and the coke can.
[Answer] No.

Here is another example:
[Question] Which object can be reached by kids more easily, the white and yellow rabbit toy can or the dark green can of beer?
[You] What is the elevation of the white and yellow rabbit toy can?
[Friend] 0.9 m.
[You] What is the elevation of the dark green can of beer?
[Friend] 0.2 m.
[Answer] Since the kids are generally shorter, it is easier for them to reach something that are lower in altitude, so it would be easier for them to reach the can of beer.

Now, given a new question, try to answer the questions by asking your friend for related visual information.
[Question]
\end{lstlisting}

By doing so, we find LLM and \algo can effectively work together to derive the correct .

\subsection{Implementation Details}

\paragraph{Semantic Filtering}
\label{appendix:data_iltering}
In the data filtering phase, we have $2$ important objectives: First, we shall filter out images that humans can hardly ask any spatial questions, such as photo of a single object before a white background. Second, since our process requires lifting a 2D image to 3D point cloud, we desire the field of view to be close to a value our monocular depth estimation model is optimized for.

To achieve the first objective, we use a pretrained CLIP model to label the candidate images, and filter out those that represent a product or an artwork. Positive CLIP labels include ``an iphone photo of an indoor scene", and ``an iphone photo of an outdoor scene", while negative labels are ``a close up shot of a single object",
        ``a product displayed in front of a white background",
        ``an artwork",
        ``a painting",
        ``a screenshot of graphics user interface",
        ``a piece of text",
        ``a sketch".

We choose ``an iphone photo'' as a prefix for positive cases to satisfy the second objective. We observe that this prefix effectively filters out data that has a wider field of view, as well as certain images with uncommon perspective ratio. 

Such design choices in data filtering ensure the images left are within the effective distribution of our expert models and qa generation.

\paragraph{2D Contexts Extraction}
As we mentioned in method section, we use a variety of off-the-shelf models to extract relavent information to synthesize our question answer data. Here we provide additional details to context extraction. After data filtering, we run a region proposal network (RPN) followed by a non-max suppression (NMS)~\cite{he2017mask}. For each object bounding box, we run a class agnostic segmentation model~\cite{kuo2019shapemask} to segment out the object. For each bounding box, we use FlexCap~\cite{anonymous2023flexcap} to sample an object-centric caption with random length between $1-6$ words. In particular, we deliberately choose to avoid traditional object detectors, as they are fixed to very coarse categories such as ``cake'', while our approach can annotate objects with fine-grained descriptions like ``cake shaped like a house'' and ``cup cake in plastic container'' for the image in Figure~\ref{fig:method}.

\paragraph{2D Context to 3D Context Lifting}
We then run the state-of-the-art metric depth detector, ZoeDepth~\cite{bhat2023zoedepth} on the image. ZoeDepth outputs metric depth (in real-world ``meters''). Combined with an fov estimation, we are able to to lift 2D images into 3D point clouds as illustrated with real-world scale as illustrated in Figure~\ref{fig:method}. 

In this point cloud processing step, outliers, or points that significantly deviate from the main group, are removed to enhance data accuracy. A clustering algorithm DBSCAN~\cite{ester1996density} groups the points based on proximity, focusing on densely populated regions and eliminating sparse, less significant points. This results in a cleaner, more structured point cloud, ideal for subsequent shape and geometry analysis where dimensions of the shapes are measured. Since we already obtained the semantic segmentation for the point cloud, we may use this information to process outliers at adaptive scales. For smaller objects, we use a smaller threshold, proportional to the along each axis. We observe that such choice effectively removes point cloud outliers while also keeping important points for smaller objects. We provide algorithm details below in Algorithm \ref{lst:outlier_removal}.

\begin{lstlisting}[basicstyle=\ttfamily\scriptsize, backgroundcolor = \color{lightgray}, keywordstyle = {\textbf}, caption={Outlier removal with DBScan}, label={lst:outlier_removal}]
Input:
points_obj: Pointcloud of object of interest
pcd: Full Pointcloud

scale = norm(points_obj.std(axis=0)) * 3.0 + 1e-6
pcd = pcd.remove_stat_outlier(neighbors=50, std=1.2)
pcd = pcd.down_sample(voxel_size=max(0.01, scale / 20))
labels = array(pcd.cluster_dbscan(
    eps=scale / 3.6, min_points=len(pcd) // 10))

\end{lstlisting}

\begin{figure*}[h]
  \centering
  \includegraphics[width=0.8\linewidth]{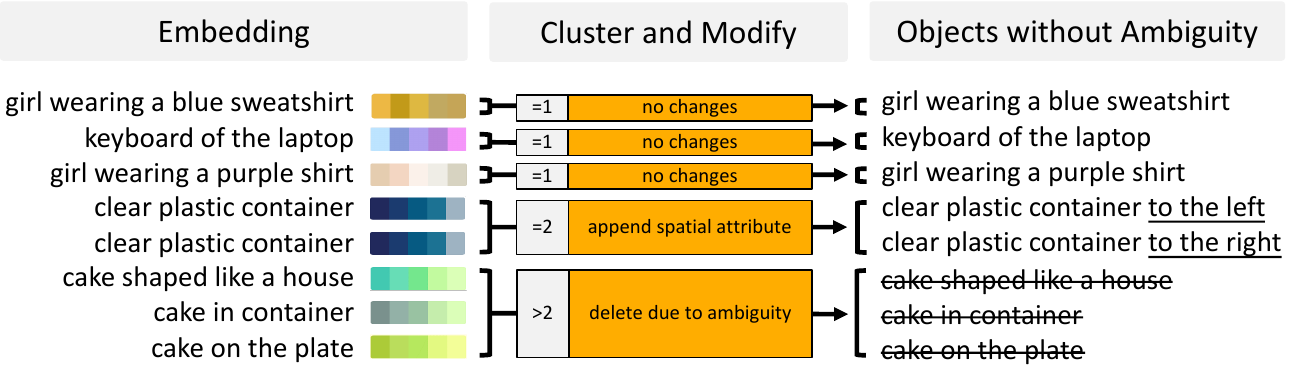}
  \caption{This is a figure illustrating ambiguity removal.}
  \label{fig:ambiguity}
\end{figure*}

\paragraph{Coordinate Canonicalization}
Now we have a 3D point cloud under metric scale. However, the point cloud is still in camera frame, which limits the information we can extract. For example, an object closer to the upper side of the image isn't necessarily further from the ground, because the camera might be pointing at the ground instead of the front. In order to solve this problem, we canonicalize the coordinate system of the pointcloud by detecting horizontal surfaces. We use a light weight segmentation model~\cite{chen2017deeplab} to segment out pixels correspond to categories like ``floor'', ``table top'', before using RANSAC to fit the biggest plane among these 3D points.

When we detect a surface defined by enough points, we canonicalize the coordinate of the point cloud by creating a new origin by projecting camera origin to the detected plane. We use the normal axis of the detected plane as z-axis and project the original z-axis of camera on the the plane as the new x-axis. By doing so, when we can detect horizontal surfaces like ground, we effectively transform the point cloud into world coordinate instead of camera coordinate. A more detailed algorithm can be found in our algorithm box \ref{lst:canonicalization}.
On the other hand, when not enough points corresponding to horizontal surfaces are detected, we flag canonicalization as failed and avoid synthesizing questions that depends on canonicalization, such as questions about elevation.

\begin{lstlisting}[basicstyle=\ttfamily\scriptsize, backgroundcolor = \color{lightgray}, keywordstyle = {\textbf}, caption={Canonicalization Algorithm}, label={lst:canonicalization}]
Input:
depth: predicted depth for each point
ground_mask: detected ground or not for each point

points_cam = unproject_to_pointcloud(depth, fov)
points = points_cam
ground_mask = ground_mask.flatten()
canonicalized = False

if ground_mask.mean() > canonicalize_threshold:
  canonicalized = True
  ground_pcd = subset(points_cam, ground_mask)
  plane, _ = ground_pcd.segment_plane(
      distance_threshold=0.05, ransac_n=3, num_iterations=1000)
  if array([0, -1, 0]) @ plane[:3] < 0:
    plane = -plane
  a, b, c, d = plane
  normal = array([a, b, c])
  ez = array([0, 0, 1])
  new_y = ez - normal @ ez * normal
  new_y = new_y / norm(new_y)
  rot = array([cross_prod(new_y, normal), new_y, normal])
  rot = array([[0, -1, 0], [1, 0, 0], [0, 0, 1]]) @ rot
  trans = array([0, 0, d])
  points_world = points_cam @ rot.T + trans[None]
  points = points_world

return points, canonicalized
\end{lstlisting}

\paragraph{Ambiguity Removal}
\label{appendix:ambiguity}
As shown in Figure ~\ref{fig:ambiguity}, we first embed all captions with CLIP encoder~\cite{radford2021learning}. This allows us to calculate a cosine distance between each caption pair. This forms a similarity matrix between all objects. If we threshold the similarity score, we can identify whether an object caption is too close to others. In many cases, we have groups of exactly two similar captions, so we can easily augment each caption by appending an differentiating clause such as ``that's more to the top of the image''. Other cases involve more than two similar captions, which we choose to remove all together to avoid ambiguity. We also remove common background objects based on CLIP similarity to categories like ``sun'' or ``sky''.

\begin{figure*}[h]
  \centering
  \includegraphics[width=0.8\linewidth]{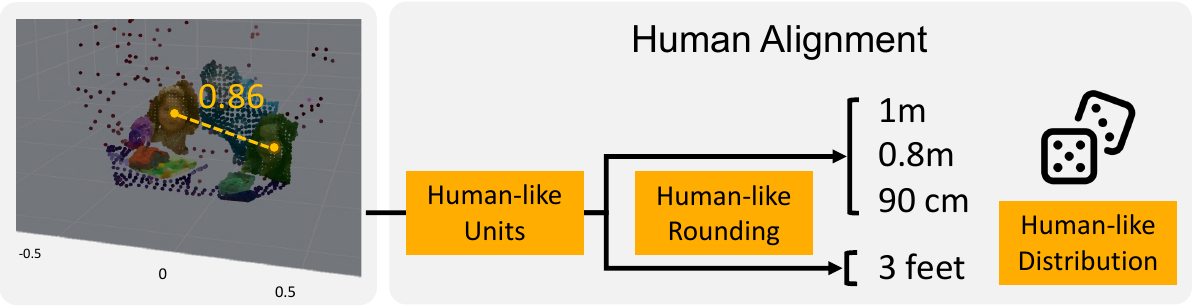}
  \caption{In human alignment, we define a set of rules that rounds with a probability that mimics the decision rules of humans.}
  \label{fig:alignment}
\end{figure*}

\paragraph{Human Alignment}
\label{appendix:alignment}
Humans rarely say a distance measure with many decimal places like $0.95$ meters. Rather, they round such distance into some thing they prefer, such as $1$ meter or half a meter. We would like our model to align with such human preferences as well. In fact, since depth estimation and fov estimation contain irreducible errors, the model should be allowed to phrase its answers with uncertainty by rounding just like humans, unless when prompted to be accurate. To this end, we post process any quantitative distance unit to align with human preferences. As illustrated in Figure \ref{fig:alignment}, we coded a decision tree to make such alignment. For example, when the estimated distance is $0.86$ meters, with $75\%$ probability we just round it to $1$ meters, while we answer $3$ feet, $90$ cm with some lower probabilities. For a distance like $23$ meters, we round it to $20$ meters with high probability as well. We also sample imperial units instead of metric units by a $20\%$ chance, with similar human-like rounding rules.

While this may align the model better with humans, at sampling time, one may want to get more accurate distance estimation. To do so, one simply need to sample multiple distance estimations and take the average. We used this in our robotic experiments to get more fine-grained values. Another way is to prompt VLM itself to keep a certain amount of digits. This can be added to data synthesis but we leave this to future work.

\paragraph{VLM Training}
We trained our multi-modal large language model with a batch size of 512 and an ADAM optimizer with learning rate of 2e-4. We trained the PaLM 2-E-S model using a mixture of the original VQA datasets in PaLM-E and our generated spatial VQA dataset, with a sampling ratio of \textit{174:2.5}. We initially train the model with a frozen vision encoder for 110k steps, which doesn't use all the data we exhausted. Therefore the data we generated is more than enough. We then, like described in the experiment section, finetune with the vision encoder either frozen or unfrozen for 70k steps till convergence.

\subsection{Question and Answer Template}
\label{appendix:vqa_template}
As we mentioned in our method section, we synthesis question and answering pairs via templates. Given a description of a pair of objects, such as ``the yellow banana'' and ``the cake in the shape of a house'', our data synthesis pipeline can effectively extract an answer based on question types. 

Here we provide a distribution of the question and answer types in Figure~\ref{fig:qa_dist_canonicalized_false} and Figure~\ref{fig:qa_dist_canonicalized_true}, followed by a description about each category.

\begin{enumerate}
    \item \textbf{left predicate} A question asking whether object A is to the left of object B from the viewer's perspective. The solution is a binary predicate true or false expressed in natural language, or a phrase expressing uncertainty.
    \item \textbf{right predicate} A question asking whether object A is to the right of object B from the viewer's perspective. The solution is a binary predicate true or false expressed in natural language, or a phrase expressing uncertainty.
    \item \textbf{above predicate} A question asking whether object A is above object B. Requires coordinate canonicalization. The solution is a binary predicate true or false expressed in natural language, or a phrase expressing uncertainty.
    \item \textbf{below predicate} A question asking whether object A is below object B. Requires coordinate canonicalization. The solution is a binary predicate true or false expressed in natural language, or a phrase expressing uncertainty.
    \item \textbf{behind predicate} A question asking whether object A behind object B from the viewer's perspective. The solution is a binary predicate true or false expressed in natural language, or a phrase expressing uncertainty.
    \item \textbf{front predicate} A question asking whether object A is in front of object B. The solution is a binary predicate true or false expressed in natural language, or a phrase expressing uncertainty.
    \item \textbf{tall predicate} A question asking whether object A is taller than object B. Requires coordinate canonicalization. The solution is a binary predicate true or false expressed in natural language, or a phrase expressing uncertainty. 
    \item \textbf{short predicate} A question asking whether object A is shorter than object B. Requires coordinate canonicalization. The solution is a binary predicate true or false expressed in natural language, or a phrase expressing uncertainty.
    \item \textbf{wide predicate} A question asking whether object A is wider than object B. The solution is a binary predicate true or false expressed in natural language, or a phrase expressing uncertainty. 
    \item \textbf{thin predicate} A question asking whether object A is thiner than object B. The solution is a binary predicate true or false expressed in natural language, or a phrase expressing uncertainty.
    \item \textbf{big predicate} A question asking whether object A is bigger than object B. Requires coordinate canonicalization. The solution is a binary predicate true or false expressed in natural language, or a phrase expressing uncertainty. 
    \item \textbf{small predicate} A question asking whether object A is smaller than object B. Requires coordinate canonicalization. The solution is a binary predicate true or false expressed in natural language, or a phrase expressing uncertainty.

    \item \textbf{left choice} A question asking which of object A and object B is more to the left from the viewer's perspective. The solution is an object name expressed in natural language, or a phrase expressing uncertainty.
    \item \textbf{right choice} A question asking which of object A and object B is more to the right from the viewer's perspective. The solution is an object name expressed in natural language, or a phrase expressing uncertainty.
    \item \textbf{above choice} A question asking which of object A and object B is more above. Requires coordinate canonicalization. The solution is an object name expressed in natural language, or a phrase expressing uncertainty.
    \item \textbf{below choice} A question asking which of object A and object B is more below. Requires coordinate canonicalization. The solution is an object name expressed in natural language, or a phrase expressing uncertainty.
    
    \begin{figure*}[h]
    \centering
    \includegraphics[width=0.8\textwidth]{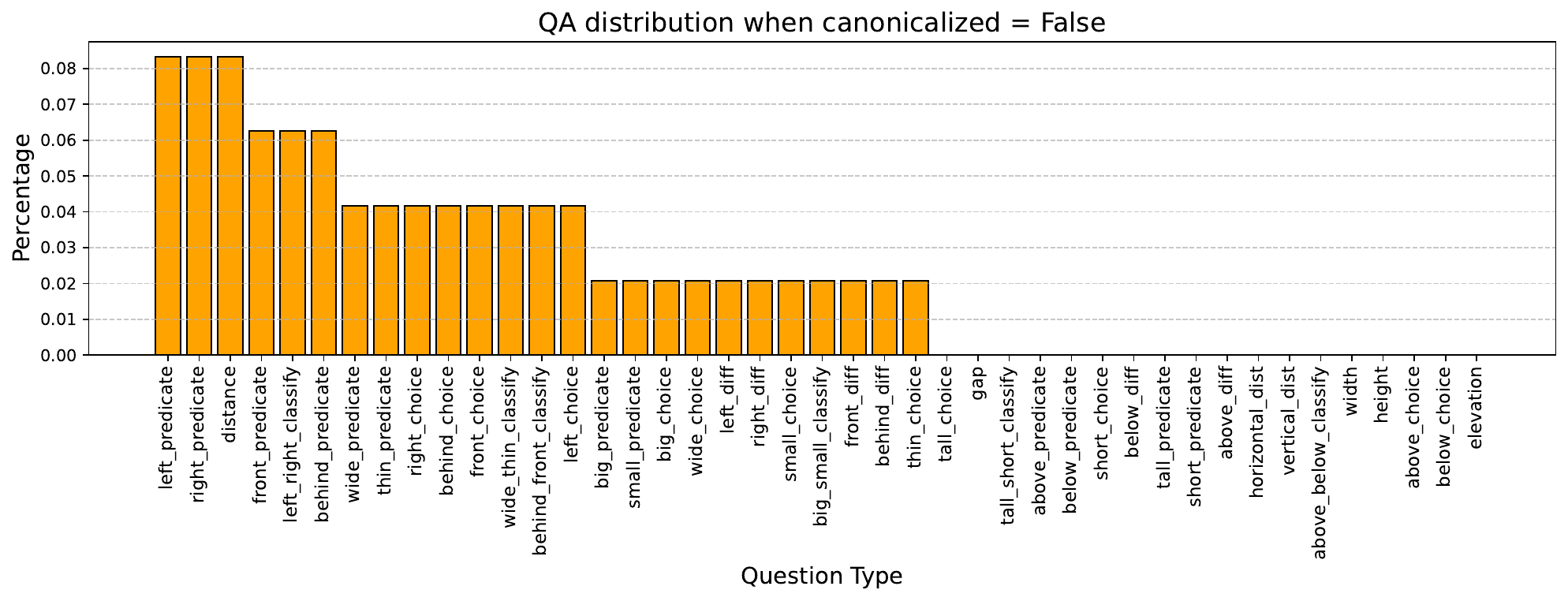}
    \caption{Distribution of generated question-answer categories when canonicalization fails.}
    \label{fig:qa_dist_canonicalized_false}
    \end{figure*}
    
    \begin{figure*}[h]
    \centering
        \includegraphics[width=0.8\textwidth]{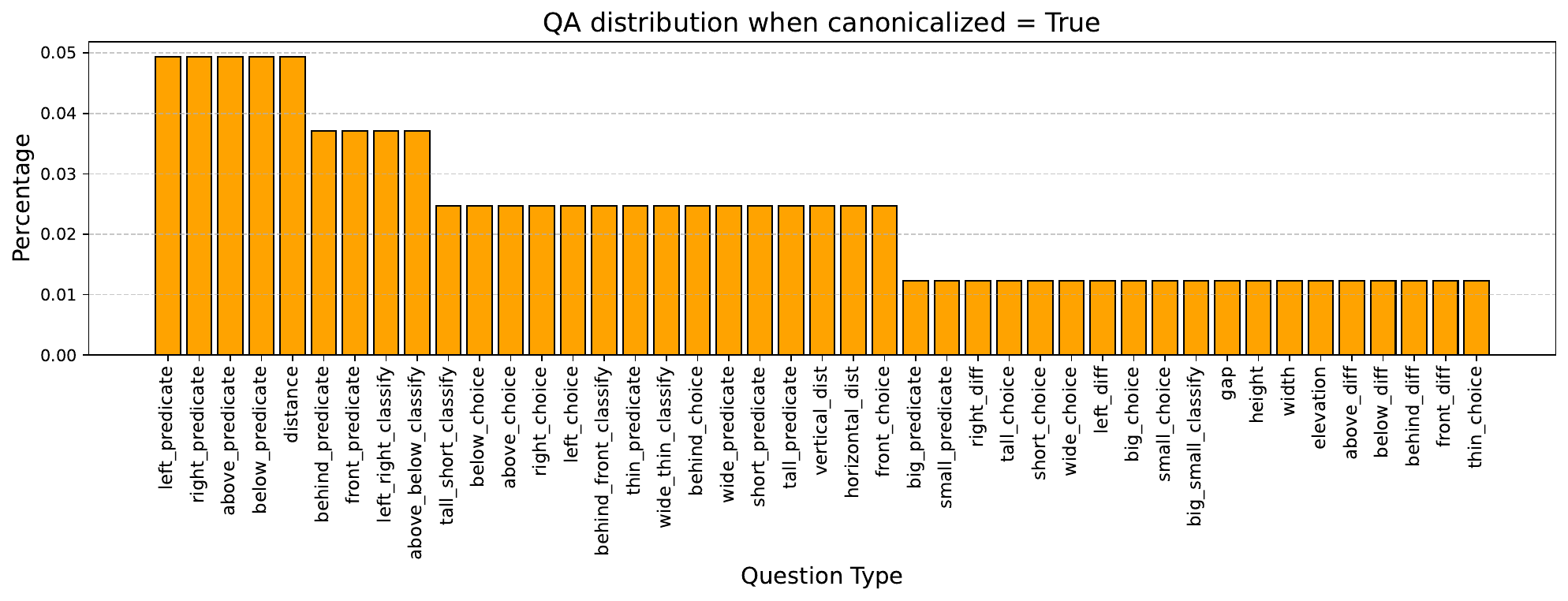}
        \caption{Distribution of generated question-answer categories when canonicalization is successful.}
        \label{fig:qa_dist_canonicalized_true}
    \end{figure*}

    \item \textbf{behind choice}  A question asking which of object A and object B is more behind. The solution is an object name expressed in natural language, or a phrase expressing uncertainty.
    \item \textbf{front choice} A question asking which of object A and object B is more to the front from the viewer's perspective. The solution is an object name expressed in natural language, or a phrase expressing uncertainty.
    \item \textbf{tall choice} A question asking which of object A and object B is taller. Requires canonicalization. The solution is an object name expressed in natural language, or a phrase expressing uncertainty.
    \item \textbf{short choice} A question asking which of object A and object B is shorter. Requires canonicalization. The solution is an object name expressed in natural language, or a phrase expressing uncertainty.
    \item \textbf{wide choice} A question asking which of object A and object B is wider. The solution is an object name expressed in natural language, or a phrase expressing uncertainty.
    \item \textbf{thin choice} A question asking which of object A and object B is thinner. The solution is an object name expressed in natural language, or a phrase expressing uncertainty.
    \item \textbf{big choice} A question asking which of object A and object B is bigger. The solution is an object name expressed in natural language, or a phrase expressing uncertainty.
    \item \textbf{small choice} A question asking which of object A and object B is smaller. The solution is an object name expressed in natural language, or a phrase expressing uncertainty.
    
    \item \textbf{left-right classify}  A question asking about the left-right comparative relationship between two objects. The solution is left-right expressed in natural language, or a phrase expressing uncertainty.
    \item \textbf{above-below classify}  A question asking about the above-below comparative relationship between two objects. Requires canonicalization. The solution is above-below expressed in natural language, or a phrase expressing uncertainty.
    \item \textbf{behind-front classify}  A question asking about the behind-front comparative relationship between two objects. The solution is behind-front expressed in natural language, or a phrase expressing uncertainty.
    \item \textbf{tall-short classify}  A question asking about the tall-short comparative relationship between two objects.  Requires canonicalization. The solution is tall-short expressed in natural language, or a phrase expressing uncertainty.
    \item \textbf{wide-thin classify}  A question asking about the wide-thin comparative relationship between two objects. The solution is wide-thin expressed in natural language, or a phrase expressing uncertainty.
    \item \textbf{big-small classify}  A question asking about the big-small comparative relationship between two objects.  Requires canonicalization. The solution is big-small expressed in natural language, or a phrase expressing uncertainty.

    \item \textbf{distance estimation}  A question asking about the distance between the center of two objects. The solution is a distance expressed in natural language, with a human-like distribution for rounding.
    \item \textbf{gap estimation}  A question asking about the gap between two objects. The solution is a distance expressed in natural language, with a human-like distribution for rounding.
    \item \textbf{height estimation}  A question asking about the height of an object. The solution is a distance expressed in natural language, with a human-like distribution for rounding. Requires canonicalization.
    \item \textbf{width estimation}  A question asking about the width of an object. The solution is a distance expressed in natural language, with a human-like distribution for rounding. 
    \item \textbf{elevation estimation}  A question asking about the elevation of an object. The solution is a distance expressed in natural language, with a human-like distribution for rounding. Requires canonicalization.
    \item \textbf{vertical distance estimation}  A question asking about the vertical distance between the center of two objects. The solution is a distance expressed in natural language, with a human-like distribution for rounding. Requires canonicalization.
    \item \textbf{horizontal distance estimation}  A question asking about the horizontal distance between the center of two objects. The solution is a distance expressed in natural language, with a human-like distribution for rounding. Requires canonicalization.
    
    \item \textbf{above difference estimation}  A question asking about the the distance between the bottom of more elevated object and the bottom of the less elevated object. The solution is a distance expressed in natural language, with a human-like distribution for rounding. Requires canonicalization.
    \item \textbf{below difference estimation}  A question asking about the distance between the bottom of less elevated object and the bottom of the more elevated object. The solution is a distance expressed in natural language, with a human-like distribution for rounding. Requires canonicalization.
    \item \textbf{behind difference estimation}  A question asking about how much an object is in behind another a distance alone the camera ray. The solution is a distance expressed in natural language, with a human-like distribution for rounding. 
    \item \textbf{front difference estimation}  A question asking about how much an object is in in front of another a distance alone the camera ray. The solution is a distance expressed in natural language, with a human-like distribution for rounding. 
    \item \textbf{left difference estimation}  A question asking about how much an object is to the left of another, from the viewer's perspective. The solution is a distance expressed in natural language, with a human-like distribution for rounding. 
    \item \textbf{right difference estimation}  A question asking about how much an object is to the right of another, from the viewer's perspective. The solution is a distance expressed in natural language, with a human-like distribution for rounding. 
\end{enumerate}

We provide a small set of question and answer pairs for generating QA data. For the full list please refer to our website.

\begin{lstlisting}[basicstyle=\ttfamily\scriptsize, backgroundcolor = \color{lightgray}, keywordstyle = {\textbf}, caption={\algo Question and Answer Template}, label={lst:template}]

OBJ_A = "[A]"
OBJ_B = "[B]"
DIST = "[X]"

distance_questions = [
    "What is the distance between [A] and [B]?",
    "How far apart are [A] and [B]?",
    "How distant is [A] from [B]?",
    "How far is [A] from [B]?",
    "How close is [A] from [B]?",
    "Could you measure the distance between [A] and [B]?",
    "Can you tell me the distance of [A] from [B]?",
    "How far away is [A] from [B]?",
    "Can you provide the distance measurement between [A] and [B]?",
    "Can you give me an estimation of the distance between [A] and [B]?",
    "Could you provide the distance between [A] and [B]?",
    "How much distance is there between [A] and [B]?",
    "Tell me the distance between [A] and [B].",
    "Give me the distance from [A] to [B].",
    "Measure the distance from [A] to [B].",
    "Measure the distance between [A] and [B].",
]


distance_answers = [
    "[X]",
    "[A] and [B] are [X] apart.",
    "[A] is [X] away from [B].",
    "A distance of [X] exists between [A] and [B].",
    "[A] is [X] from [B].",
    "[A] and [B] are [X] apart from each other.",
    "They are [X] apart.",
    "The distance of [A] from [B] is [X].",
]


vertical_distance_questions = [
    "What is the vertical distance between [A] and [B]?",
    "How far apart are [A] and [B] vertically?",
    "How distant is [A] from [B] vertically?",
    "How far is [A] from [B] vertically?",
    "Could you measure the vertical distance between [A] and [B]?",
    "Can you tell me the vertical distance between [A] and [B]?",
    "How far away is [A] from [B] vertically?",
    (
        "Can you provide the measurement of the vertical distance between [A]"
        " and [B]?"
    ),
    "Estimate the vertical distance between [A] and [B].",
    "Could you provide the vertical distance between [A] and [B]?",
    "How much distance is there between [A] and [B] vertically?",
    "Tell me the distance between [A] and [B] vertically.",
    "Give me the vertical distance from [A] to [B].",
    "Measure the vertical distance from [A] to [B].",
    "Measure the distance between [A] and [B] vertically.",
]


vertical_distance_answers = [
    "[X]",
    "[A] and [B] are [X] apart vertically.",
    "[A] is [X] away from [B] vertically.",
    "A vertical distance of [X] exists between [A] and [B].",
    "[A] is [X] from [B] vertically.",
    "[A] and [B] are [X] apart vertically from each other.",
    "Vertically, They are [X] apart.",
    "The vertical distance of [A] from [B] is [X].",
    "They are [X] apart.",
    "It's approximately [X]."
]


horizontal_distance_questions = [
    "What is the horizontal distance between [A] and [B]?",
    "How far apart are [A] and [B] horizontally?",
    "How distant is [A] from [B] horizontally?",
    "How far is [A] from [B] horizontally?",
    "Could you measure the horizontal distance between [A] and [B]?",
    "Can you tell me the horizontal distance of [A] from [B]?",
    "How far away is [A] from [B] horizontally?",
    (
        "Can you provide the measurement of the horizontal distance between [A]"
        " and [B]?"
    ),
    (
        "Can you give me an estimation of the horizontal distance between [A]"
        " and [B]?"
    ),
    "Could you provide the horizontal distance between [A] and [B]?",
    "How much distance is there between [A] and [B] horizontally?",
    "Tell me the distance between [A] and [B] horizontally.",
    "Give me the horizontal distance from [A] to [B].",
    "Vertial gap between [A] and [B].",
    "Measure the horizontal distance from [A] to [B].",
    "Measure the distance between [A] and [B] horizontally.",
]


horizontal_distance_answers = [
    "[X]",
    "[A] and [B] are [X] apart horizontally.",
    "[A] is [X] away from [B] horizontally.",
    "A horizontal distance of [X] exists between [A] and [B].",
    "[A] is [X] from [B] horizontally.",
    "[A] and [B] are [X] apart horizontally from each other.",
    "Horizontally, They are [X] apart.",
    "The horizontal distance of [A] from [B] is [X].",
    "They are [X] apart.",
    "It's approximately [X]."
]

width_questions = [
    "Measure the width of [A].",
    "Determine the horizontal dimensions of [A].",
    "Find out how wide [A] is.",
    "What is the width of [A]?",
    "How wide is [A]?",
    "What are the dimensions of [A] in terms of width?",
    "Could you tell me the horizontal size of [A]?",
    "What is the approximate width of [A]?",
    "How wide is [A]?",
    "How much space does [A] occupy horizontally?",
    "How big is [A]?",
    "How big is [A] in terms of width?",
    "What's the radius of [A]?"
]

width_answers = [
    "[X]",
    "The width of [A] is [X].",
    "[A] is [X] wide.",
    "[A] is [X] in width.",
    "It's [X]."
]

behind_predicate_questions = [
    "Is [A] behind [B]?",
    "Is the position of [A] more distant than that of [B]?",
    "Does [A] lie behind [B]?",
    "Is [A] positioned behind [B]?",
    "Is [A] further to camera compared to [B]?",
    "Does [A] come behind [B]?",
    "Is [A] positioned at the back of [B]?",
    "Is [A] further to the viewer compared to [B]?",
]


behind_true = [
    "Yes.",
    "Yes, it is.",
    "Yes, it's behind [B].",
    "That's True.",
    "Yes, [A] is further from the viewer.",
    "Yes, [A] is behind [B]."
]


behind_false = [
    "No.",
    "No, it is not.",
    "No, it's in front of [B].",
    "That's False.",
    "No, [A] is closer to the viewer.",
    "No, [B] is in front of [A]."
]


front_predicate_questions = [
    "Is [A] in front of [B]?",
    "Is the position of [A] less distant than that of [B]?",
    "Does [A] lie in front of [B]?",
    "Is [A] positioned in front of [B]?",
    "Is [A] closer to camera compared to [B]?",
    "Does [A] come in front of [B]?",
    "Is [A] positioned before [B]?",
    "Is [A] closer to the viewer compared to [B]?",
]


front_true = [
    "Yes.",
    "Yes, it is.",
    "Yes, it's in front of [B].",
    "That's True.",
    "Yes, [A] is closer to the viewer.",
    "Yes, [A] is in front of [B]."
]


front_false = [
    "No.",
    "No, it is not.",
    "No, it's behind [B].",
    "That's False.",
    "No, [A] is further to the viewer.",
    "No, [B] is behind [A]."
]
\end{lstlisting}

\end{document}